\documentclass[preprint,12pt]{elsarticle}

\usepackage{amscd,amssymb,graphics,lineno}
\usepackage{mathrsfs} 
\usepackage{amsmath}

\newtheorem{theorem}{Theorem}[section]
 
\newtheorem{lemma}[theorem]{Lemma}

\newtheorem{conjecture}[theorem]{Conjecture}
\newdefinition{definition}[theorem]{Definition}
\newdefinition{remark}[theorem]{Remark}
\newdefinition{example}[theorem]{Example}
\newproof{proof}{Proof} 
\newproof{pot1}{On the proof of Theorem \ref{t:stone2}}
\newproof{pot2}{Proof of Theorem \ref{th:main}}

\def\norm#1{\left\Vert#1\right\Vert}

\def\I {{\mathbb I}}

\def\E{{\mathbb{E}}}

\def\R{{\mathbb R}}
\def\e{\varepsilon}

\def\ve{\varepsilon}

\def\poly{{\mathrm{poly}}\,}

\journal{Computers \& Mathematics with Applications}

\begin{document}

\begin{frontmatter}

\title{Is the $k$-NN classifier in high dimensions affected by the curse of dimensionality?}

\author{Vladimir Pestov}
\ead{vpest283@uottawa.ca}
\address{Department of Mathematics and Statistics, 
University of Ottawa, 585 King Edward Ave., Ottawa, Ontario, Canada K1N 6N5\fnref{label3}}
\fntext[label3]{{\em Tel.} 613-562-5800 ext 3523, {\em fax} 613-562-5776}
\begin{abstract}
There is an increasing body of evidence suggesting that exact nearest neighbour search in high-dimensional spaces is affected by the curse of dimensionality at a fundamental level. Does it necessarily mean that the same is true for $k$ nearest neighbours based learning algorithms such as the $k$-NN classifier? We analyse this question at a number of levels and show that the answer is different at each of them. As our first main observation, we show the consistency of a $k$ approximate nearest neighbour classifier. However, the performance of the classifier in very high dimensions is provably unstable. As our second main observation, we point out that the existing model for statistical learning is oblivious of dimension of the domain and so every learning problem admits a universally consistent deterministic reduction to the one-dimensional case by means of a Borel isomorphism.
\end{abstract}

\end{frontmatter}

\section{Introduction}

Local learning algorithms such as $k$-NN classification or $k$-NN regression occupy an important place in statistical learning theory, further enhanced by 
a surprising recent result \cite{ZR} stating that every consistent learning algorithm, $\mathcal L$, in the Euclidean space is ``localizable'' in a suitable sense. Suppose that we show to $\mathcal L$ only data in the $r$-ball around each point $x$,
\[x\mapsto {\mathcal L}(B_r(x)\cap \sigma)(x) = f(x).\]

\begin{figure}[ht]
\begin{center}
  \scalebox{0.25}[0.25]{\includegraphics{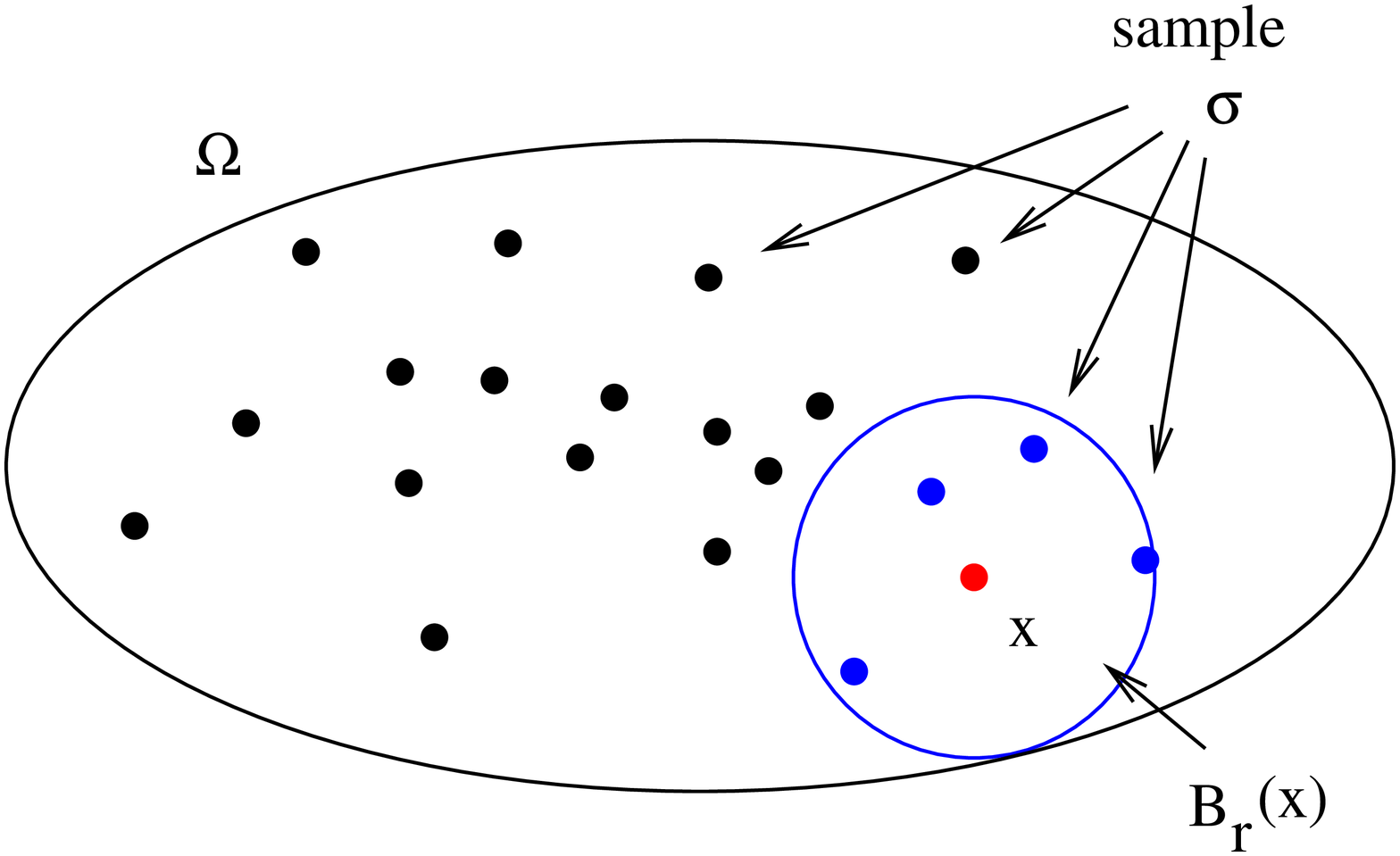}} 
 \caption{Towards the notion of localizability.}
   \label{fig:loc}
\end{center}
\end{figure}

Now smooth the resulting predictor function $f$ over the $r\cdot q$-neighbourhood of each $x$:
\[x\mapsto \E(f\vert_{B_{rq}(x)})=g(x).\]
Zakai and Ritov have shown that if $r,q\downarrow 0$ sufficiently slowly,
then the resulting ``localized'' predictor $g$ is consistent. 

It is therefore of obvious interest to examine the question of performance of local learning algorithms in high dimensional domains. Are they {\em provably} affected by the curse of dimensionality? In this article, we will concentrate on the classical $k$-NN classifier. The question turns out to be many-layered, and the answer is different at every layer that we peel back.  

Perhaps the most basic consideration is that in order to run the $k$-NN classifier, one needs to be able to efficiently retrieve $k$ nearest neighbours to every input point of the domain. For smaller datasets, this problem is solved by means of a complete sequential scan of data. However, for larger datasets this becomes impracticable, and considerable efforts of the data engineering community go towards designing various {\em indexing schemes} assuring faster similarity search \cite{samet,ZADB:06}. 

In spite of all the progress in the area,
there is a considerable body of evidence in support of the so-called {\em curse of dimensionality conjecture} \cite{indyk:04} affecting the exact deterministic nearest neighbour search in high dimensional spaces. Recall that the {\em Hamming cube of rank $d$}, $\{0,1\}^d$, is the collection of all $n$-bit binary strings equipped with the {\em Hamming distance} counting the number of bits where two strings differ:
\[d(\sigma,\tau)=\sharp\{i\colon \sigma_i\neq\tau_i\}.\]
Recall also that $\omega(1)$ denotes the class of integer sequences $n_d$ that go to infinity, $O(1)$ denotes the class of all bounded sequences,
and $o(d)$ denotes the class of integer sequences $n_d$ that are infinitely small with regard to $d$, that is, $n_d/d\to 0$ as $d\to\infty$. For example, the notation $n\in d^{\omega(1)}$ means that $n$ grows faster than any  finite power of $d$, while $n\in 2^{o(d)}$ means $n$ grows slower than the $d$-th power of any number $a>1$.

In its simplest form, the conjecture states that a dataset $X$ with $n$ points in the $d$-dimensional Hamming cube $\{0,1\}^d$, where $n\in d^{\omega(1)}\cap 2^{o(d)}$, does not in general admit a data structure of size $n^{O(1)}$ (that is, polynomial in $n$) which supports exact deterministic similarity search in $X$ in time $d^{O(1)}$ (i.e., polynomial in $d$).

Even though the conjecture remains unproven in general, it has been established for some specific indexing schemes \cite{pestov:12}. Should the conjecture be proved, the $k$-NN algorithm will be affected by the curse of dimensionality simply because of a theoretical impossibility to retrieve the $k$ nearest neighbours in time polynomial in the dimension $d$ of the domain $\Omega$ without the need to store a prohibitive amount of data (superpolynomial in the size $n$ of the actual dataset).

However, it turns out that the exact nearest neighbour search is not indispensable. Approximate nearest neighbour (ANN) search \cite{IM,KOR:00} is known to admit more efficient indexing schemes than exact NN search and approximate nearest neighbours can be substitued in a classifier in place of exact ones. 
As our first main result, we propose a new local classification algorithm based on $k$ approximate nearest neighbours ($k$-ANN classifier), and prove its consistency under the assumption of absolute continuity of the data distribution. Theorem \ref{t:kann} is a (partial) extension of
the classical Stone consistency theorem \cite{stone:77}. 

At the same time, we observe that in the asymptotic setting of high dimensions, $d\to\infty$, the $k$-ANN classifier is affected by what may be regarded a variant of Hughes's phenomenon \cite{hughes:68}. Namely, the number of datapoints required to maintain the consistency of the algorithm must provably grow exponentially with the dimension of the domain, which assumption is of course unrealistic. Thus, at least in an artificial theoretical setting of data sampled randomly from high dimensional distributions, switching to the ANN classifier does not lift the curse of dimensionality.

Here comes the second main result of the article which, in spite of its simplicity, is quite interesting (Theorem \ref{th:main}): Stone's theorem is insensitive to the Euclidean structure on the domain as long as the underlying Borel structure remains invariant. This allows for a very simple ``Borel isomorphic data reduction'' to the one-dimensional case, after which the $k$-NN algorithm still remains universally consistent. Moreover, such a consistent reduction to the one-dimensional real case applies to functional data classification in infinite-dimensional spaces, in fact in any separable metric space. 

The {\em Borel structure} is a subject of study of the descriptive set theory \cite{kechris}. It is a derivative of the usual topology of the Euclidean or infinite-dimensional Banach space, and a considerably coarser structure than the topology, preserving significantly less information. As a result, {\em Borel isomorphisms} between the domains --- that is, bijections preserving the Borel structure, possibly discontinuous at every point --- are very numerous and easy to come by. Every Euclidean space $\R^n$, in fact every infinite-dimensional Banach or even Fr\'echet linear space is Borel-isomorphic to the real line, and the corresponding isomorphisms can be easily managed at an algorithmic level and implemented in code. While the Borel structure is widely used in various parts of pure mathematics, including foundations of theoretical probability, we are unaware of examples of it being employed for the purposes of algorithmic data analysis.

The practical significance of this observation still remains to be seen (it has only been tested on a few toy datasets from the UCI repository, with encouraging results), but
on a theoretical level it brings up the problem of what is ``dimension'' in the context of statistical learning. We conclude the paper with a small discussion.

The presentation of results in our paper follows the order in the Introduction, and is preceded by a reminder of the standard model of statistical learning and the Stone consistency theorem.

\section{Stone's theorem\label{s:stone}}

Here we recall a fundamental result which serves as a theoretical justification for the $k$-NN classifier. 

The {\em domain} is, in the case of main interest for us, a $d$-dimensional Euclidean space, $\Omega=\R^d$. However, it can be any complete separable metric space. The {\em Borel structure} on $\Omega$ is the smallest family, $\mathscr B$, of subsets of $\Omega$ which contains all open balls and is closed under countable unions and complements. A function $f\colon\Omega\to\R$ is {\em Borel measurable} if the inverse image of every interval $(a,b)$ (equivalently, of every Borel subset of the real line) under $f$ is a Borel subset of $\Omega$. (For a more detailed discussion, see Subs. \ref{ss:borel}.)
A {\em Borel probability measure} $\mu$ on $\Omega$ is a countably-additive function on $\mathscr B$ with values in the interval $[0,1]$, satisfying $\mu(\Omega)=1$. 

Data pairs $(x,y)$, where $x\in\Omega$ and $y\in\{0,1\}$, follow an unknown probability distribution $\mu$ (a Borel probability measure on $\Omega\times\{0,1\}$). 
Denote $L(\Omega,\{0,1\})$ the collection of all Borel measurable binary functions on the domain.
Given such a function $f\colon\Omega\to\{0,1\}$ (a classifier), the  
{\em misclassification error} is defined by 
\[{\mathrm{err}}_{\mu}(f) = \mu\{(x,y)\in \Omega\times\{0,1\}\colon f(x) \neq y\}.\]
The {\em Bayes error} is the infimal misclassification error over all possible classifiers: 
\[\ell^{\ast}=\ell^{\ast}(\mu)=\inf_{f}{\mathrm{err}}_{\mu}(f).\]
A {\em learning rule} is a family ${\mathcal L}=\left({\mathcal L}_n\right)_{n=1}^{\infty}$, where
\[{\mathcal L}_n\colon \Omega^n\times \{0,1\}^n \to L(\Omega,\{0,1\}),~~n=1,2,\ldots
\]
and the associated evalution maps
\[\Omega^n\times \{0,1\}^n\times \Omega \ni (\sigma,x)\mapsto {\mathcal L}_n(x,y)(z) \in \{0,1\}\]
are Borel.
Here $\sigma=(x_1,\ldots,x_n,y_1,\ldots,y_n)$ is a labelled learning sample.

For example, the $k$-NN classifier is defined by selecting the value 
${\mathcal L}_n(\sigma)(x)$ in $\{0,1\}$ by the majority vote among the values of $y$ corresponding to the $k=k_n$ nearest neighbours of $x$ in the learning sample $\sigma$. For even $k$, ties may occur, which are broken with the help of random orders on the neighbours.

Data is modelled by a sequence of independent identically distributed random elements $(X_n,Y_n)$ of $\Omega\times\{0,1\}$. Denote $\bar x$ a sample path. Then the learning rule $\mathcal L_n$ only gets to see the first $n$ labelled coordinates of $\bar x$.
A learning rule $\mathcal L$ is {\em consistent} if ${\mathrm{err}}_{\mu}{\mathcal L}_n(\bar x)\to \ell^{\ast}$ in probability as $n\to\infty$. If the convergence occurs almost surely, then $\mathcal L$ is said to be {\em strongly consistent}.
Finally, $\mathcal L$ is {\em universally consistent} if it is consistent under every probability measure $\mu$. Strong universal consistency is defined in a similar way.

\begin{theorem}[Stone \cite{stone:77}] 
  Let $k=k_n\to\infty$ and $k_n/n\to 0$. Then the $k$-NN classification algorithm in $\R^d$ (with regard to the Euclidean distance) is universally consistent. 
\end{theorem}

The conclusion was subsequently strengthened to strong universal consistency, cf. Chapter 11 in \cite{DGL} and historic references.

Stone's theorem fails in more general metric spaces, even in an infinite-dimensional Hilbert space $\ell^2$. One can construct a deterministic concept in $\ell^2$ not learned by the $k$-NN classifier over a gaussian distribution 
(cf. an example in \cite{CG}, pp. 351--352, based on a contruction of Preiss \cite{preiss}).

An alternative proof of the consistency of the $k$-NN classifier, based on the Lebesgue density theorem for the Euclidean space, was given in \cite{devroye}, and in \cite{CG} it was further shown that the $k$-NN classifier is universally consistent in every metric space satisfying the Lebesgue--Besikovitch density theorem. Such metric spaces have been completely characterized by Preiss \cite{preiss:83} (they are the so-called sigma-finite dimensional metric spaces, cf. also \cite{AQ}). It would be quite interesting to give a formal proof that the universal consistency of the $k$-NN classifier in a metric space is equivalent to the validity of the Lebesgue--Besikovitch density theorem, and further to modify the original proof of Stone to make it work for every sigma-finite dimensional metric space. This would in particular lead to a new proof of the density theorem of real analysis using tools of statistical learning theory. 

Among the factors affecting the performance of the $k$-NN classifier in a high-dimensional space, the need to retrieve $k$ nearest neighbours of an input datapoint in an effective and efficient way is most apparent, and we will proceed to it now.

\section{Exact similarity search}

Let $(\Omega,\rho)$ be a metric space, and $X\subseteq \Omega$ a finite subset (dataset). The triple $W=(\Omega,\rho,X)$ is a {\em similarity workload.} The {\em $k$-nearest neighbour query} is: given $q\in\Omega$, return $k$ nearest neighbourhs to $k$ in $X$. In practice, it is often reduced by means of binary search to a sequence of {\em $\e$-range similarity queries:} given $q\in\Omega$ and $\e>0$, return all $x\in B_\e(q)\cap X$, where $B_\e(q)$ denotes the $\e$-ball around $q$. See \cite{chavez:01}.

An {\em access method} for a workload $W$ is an algorithm that correctly answers every range query. Principal examples of access methods are {\em indexing schemes}, in particular {\em hierarchical tree-based indexing schemes}. 
One popular version of such a scheme as the $M${\em -tree} \cite{CPZ:97}. For varying discussions of indexing schemes for similarity search in metric spaces, see \cite{samet,ZADB:06,chavez:01,PeSt06}.

The curse of dimensionality for access methods into high-dimensional domains is a well-known phenomenon among the practitioners, even if it is hard to pinpoint a well-documented reference (see however \cite{WSB}). At a theoretical level, the following open ``curse of dimensionality conjecture'' sums up a rather commonly held belief in the curse of dimensionality being inherent in high dimensional data.

\begin{conjecture}
[cf. \cite{indyk:04}]
Let $X\subseteq\{0,1\}^d$ be a data\-set with $n$ points, where the Hamming
cube $\{0,1\}^d$ is equipped with the Hamming ($\ell^1$) distance:
\[d(x,y) = \sharp \{i\colon x_i\neq y_i\}.\]
Suppose $d=n^{o(1)}$, but $d=\omega(\log n)$. (That is, the number of points in $X$ has intermediate growth with regard to the dimension $d$: it is  superpolynomial in $d$, yet subexponential.)
Then any data structure for exact nearest neighbour search in $X$,
with $d^{O(1)}$ query time, must use $n^{\omega(1)}$ space within the {\em cell probe model} of computation.
\end{conjecture}

For the cell probe model of computation see \cite{miltersen}; in the context of indexing schemes it is briefly discussed in \cite{pestov:12}. The best lower bound presently known for polynomial space data structures is $\Omega(d/\log n)$ \cite{barkol:00}. (See also \cite{PT} for some later improvements.)
Rigorous lower bounds superpolynomial in $d$ have been established for a number of concrete indexing schemes, notably the pivot-based schemes \cite{VolPest09,pestov:12} and the metric trees \cite{pestov:12b}. 

\section{Approximate similarity search}

The $(c,\e)$-{\em approximate nearest neighbour search} problem \cite{indyk:04}  is stated thus: given $\ve$ and $c>0$, for a $q\in\Omega$, if $\ve_{NN}(q)\leq \ve$, then return a datapoint $x\in X$ at a distance $d(q,x)<(1+c)\ve$. (Here $\ve_{NN}(q)$ denotes the distance from $q$ to the nearest neighbour in $X$.)

\begin{figure}[ht]
\begin{center}
  \scalebox{0.25}[0.25]{\includegraphics{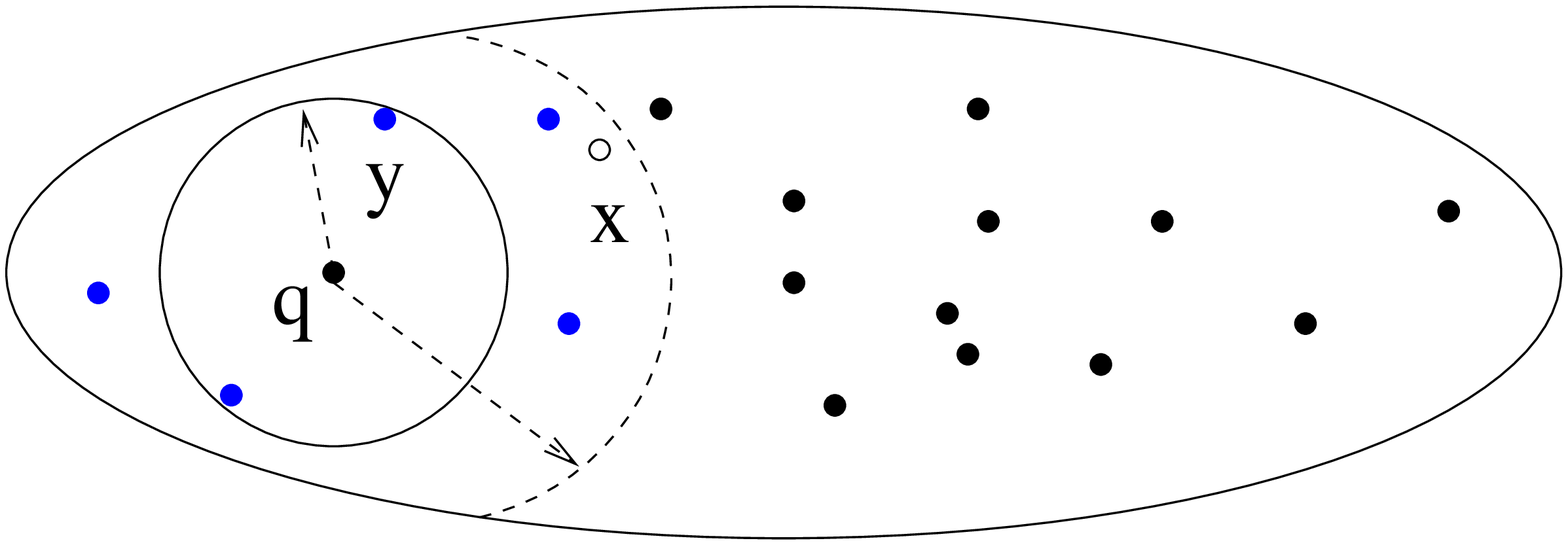}} 
  \caption{$(c,\e)$-ANN search.}
  \label{fig:ann}
\end{center}
\end{figure}

The known indexing schemes for approximate nearest neighbour (ANN) search \cite{IM,KOR:00,samet,ZADB:06} are more efficient than those for exact NN search. 

In order to be used for classification, the $(c,\e)$-ANN problem has to be modified in the following way. The {\em $(k,c)$ approximate nearest neighbours} ({\em $k$-ANN}) problem says: given $q$ and $c>0$, return $k$ datapoints contained within the distance $(1+c)\ve_{k\mbox{\tiny -NN}}$ of the query point. Here $\ve_{k\mbox{\tiny -NN}}(q)$ is the smallest radius of a ball around $q$ containing $k$ datapoints.

\begin{figure}[ht]
\begin{center}
  \scalebox{0.25}[0.2]{\includegraphics{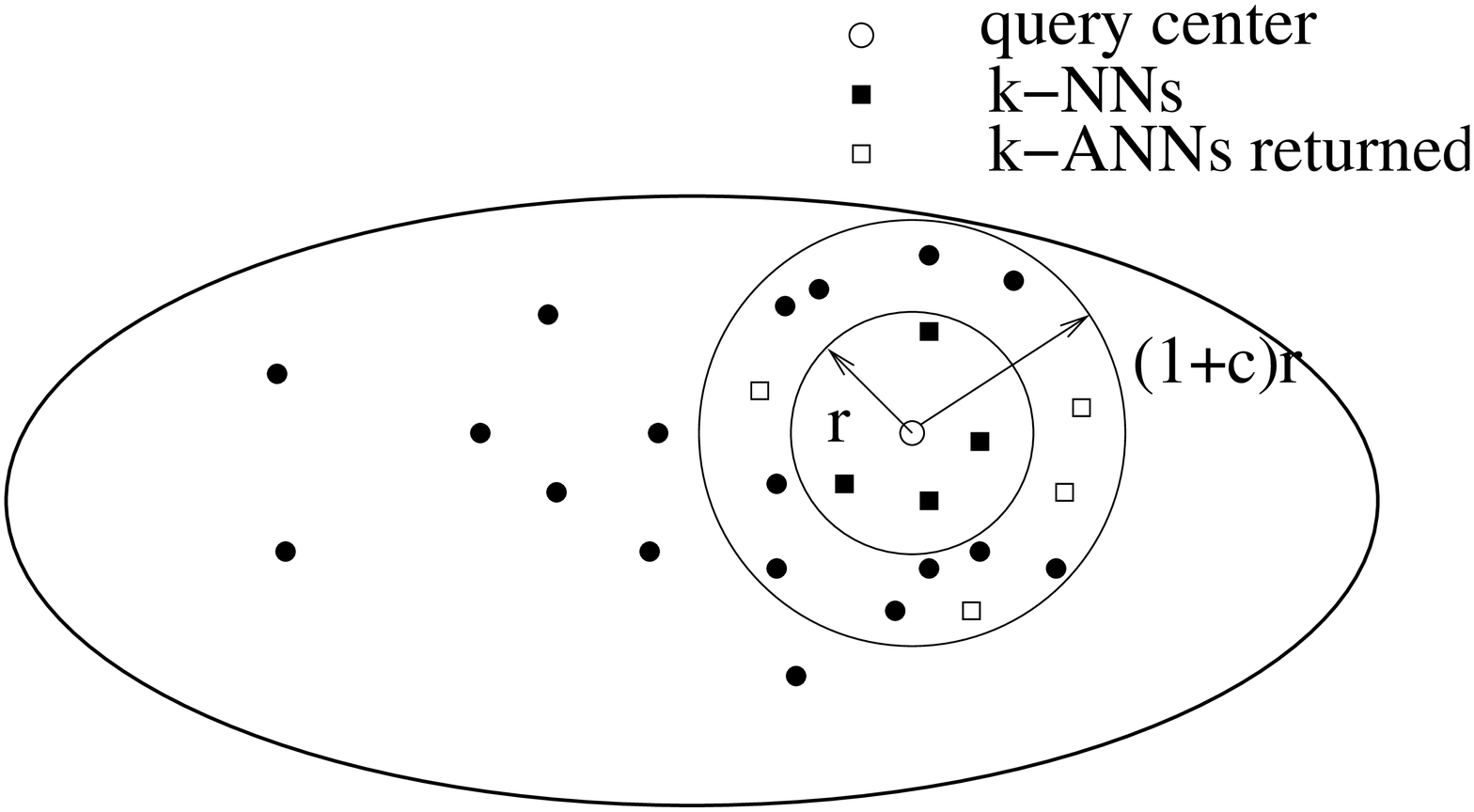}} 
  \caption{$k$-ANN search.}
  \label{fig:k-ann}
\end{center}
\end{figure}

The indexing schemes based on random projections, random matrices, or locality sensitive hashing can be adapted to answer the $(k,c)$-ANN query. 

As an example, let us consider the scheme originally developed in \cite{KOR:00} and reformulated in \cite{vempala}, Section 7.2 using random binary matrices. The core of the approach is an indexing scheme into the Hamming cube $\{0,1\}^d$, which is afterwards converted into an indexing scheme for $\R^d$ by discretization. Let $0<\e<1$, and denote $X\subseteq \{0,1\}^d$ the dataset with $n$ points.

Fix a {\em range} $\ell=1,2,\ldots,d$.
The scheme for the range $\ell$ consists of a family ${\mathscr H}$ of mappings from $\{0,1\}^d$ onto a cube of a smaller dimension $k=O(\ve^{-2}\lg_2n)$, with the following property. If $q\in \{0,1\}^d$ and a random $h\in {\mathscr H}$ is chosen (with regard to a certain probability distribution), then with a constant confidence $1-\delta$ the mapping $h$ preverves distances in the set $X\cup\{q\}$ on the scale $\ell/2$, $\ell=1,2,\ldots,d$, to within an additive error $\ell\e$, and on a larger scale --- away from it. 
In the scheme under consideration, the map $h$ is a multiplication on the right by a $d\times k$ matrix with random i.i.d. Bernoulli entries assuming values $1$ and $0$ with probabilities $1/\ell$ and $1-1/\ell$, respectively. (The operations are carried $\mathrm{mod}\, 2$.) 
The target cube only contains $2^{O(\ve^{-2}\lg_2n)}=\poly(n)$ points, and is indexed to efficiently answer a nearest neighbour query via hashing. 
The indexing scheme consists of a sufficiently large family of such functions $h$ for every possible range $\ell$. 

If we are now interested in an $(1+\ve)$-approximate nearest neighbour query, a binary search in $\ell$ finds the smallest range so that a randomly chosen $h$ only returns one nearest neighbour to $h(q)$ at a distance $\ell$. This neighbour is of the form $h(x)$, $x\in X$; the point $x$ is returned. With confidence $1-\delta$, this is a $(1+\e)$-approximate nearest neighbour of $q$ in the original Hamming cube.

If we want to increase the confidence, the algorithm is run repeatedly, and among the obtained points $x_1,x_2,\ldots$ the nearest one to $q$ is returned. 
Building the scheme takes time polynomial in $n$ and the algorithm answers {\em every} query $q$ with high confidence in time $O(d\,\mathrm{poly}\,\log (dn))$.

A modification for the $k$-ANN problem is now obvious. First, a binary search in $\ell$ determines the smallest value $\ell$ such that for the corresponding randomly chosen $h$ the $\ell$-ball around $h(q)$ contains at least $k$ datapoints. Then $k$ nearest neighbourhs, $h(x_1)$, $\ldots$, $h(x_k)$ to $h(q)$ in the cube of small dimension (image of $h$) are retrieved, and the corresponding original points $x_1,x_2,\ldots,x_k$ returned. With a constant confidence $1-\delta$, they will be $(k,\e)$ approximate nearest neighbours of $q$. Again, in order to make the confidence as high as desired, the procedure is repeated as many times as necessary, all returned points are put in a bucket, and the $k$ nearest neighbours to $q$ among them are returned. The only change in the indexing scheme is that the hash table now stores $k$ nearest neighbours instead of one. The running time of the algorithm is now $O(dk\,\mathrm{poly}\,\log (dn))$.

\section{The $k$ approximate nearest neighbour classifier}  

This section contains the first of two main new results reported in the article: an extension of the classical Stone consistency theorem \cite{stone:77} to an approximate nearest neighbour-based classifier. 

\subsection{Definition and statement of result}
Fix a $c>0$. The value of the $k$-ANN classifier (more exactly, $(k,c)$-ANN classifier) at a point $x$ is determined by the majority vote among $(k,c)$-approximate nearest neighbourhs of $x$, as returned by an indexing scheme. 

\begin{theorem} 
\label{t:kann}
Suppose the underlying data distribution $\mu$ on $\R^d\times \{0,1\}$ has density (that is, is absolutely continuous with regard to the Lebesgue measure). Let $c>0$ be fixed, and let 
\[k\in \omega(\log n)\cap o(n).\]
Then the $(k,c)$-ANN classifier is consistent. \qed
\end{theorem}

\begin{remark}
Notice that no assumption is made about the nature of the algorithm for answering $(k,c)$-ANN queries. The task can even be entrusted to an adversary who is aware of the underlying distribution $\mu$: this will not affect the consistency, though possibly slow down the rate of convergence. 
\end{remark}

\begin{remark}
  The assumption of absolute continuity of the unknown distribution $\mu$ allows us to avoid dealing with ties in the proof below. For the moment, we do not know whether this assumption can be dropped. 
\end{remark}

\begin{remark}
  We also do not know how essential is the assumption that $k$ grows strictly faster than the logarithm of $n$.
\end{remark}

\subsection{A variation on Stone's theorem}
Here is a slightly strengthened version of Stone's theorem (\cite{DGL}, Theorem 6.3).
 
\begin{theorem}
\label{t:stone2}
Let $\mu$ be a probability measure on $\R^n\times\{0,1\}$ with regard to which the datapoints are drawn as i.i.d. random variables.
Suppose that $W_{n,i} = W_{n,i}(x,X_1,X_2,\ldots,X_n)$ are data-dependent weights (random measurable functions on $\R^d$) which are nonnegative, sum up to one,
\[\sum_{i=1}^n W_{n,i}(x) = 1,\]
and satisfy the properties:
\begin{enumerate}
\item[(i)] For some $c>0$ and every Borel subset $A\subseteq\R^d$, 

\[\limsup_{n\to\infty} \E\left\{\sum_{i=1}^n W_{n,i}(x)\chi_A(X_i)\right\}\leq c\mu(A).\]
\item[(ii)] For all $a>0$,
\[\lim_{n\to\infty}\E\left\{ \sum_{i=1}^n W_{n,i}(x)I_{\norm{X_i-X}>a}\right\}=0.\]
\item[(iii)] \[\lim_{n\to\infty}\E\left\{\max_{1\leq i\leq n} W_{n,i}(x)\right\}=0.\]
\end{enumerate}
Define the classification rule $g_n$ based on the majority vote among all the values $Y_i$ each given the $W_{n,i}(x)$ share of total vote. If $\mu$ has density, then the rule $g_n$ is consistent. 
\end{theorem}

\begin{remark}
By approximating a bounded measurable function with simple functions, the condition $(i)$ is seen to be equivalent to
\begin{enumerate}
\item[(i$^\prime$)]
There is a constant $c>0$ such that
for every bounded measurable function on $\R^d$ with values in the interval, \[\limsup_{n\to\infty} \E_{\mu}\left\{\sum_{i=1}^n W_{n,i}(x)f(X_i)\right\}\leq c\E_{\mu}(f).\]
\end{enumerate}
\end{remark}

\begin{pot1}
It follows the proof of Stone's Theorem (Theorem 6.3 as presented in \cite{DGL} on pages 98--100) practically word for word. 

Notice that the conditions $(ii)$ and $(iii)$ are the same, it is only the condition $(i)$ that has been relaxed. The condition $(i)$ is only used in the proof once, to obtain the last inequality in the chain of inequalites at the end of page 99.

The functions $\eta$ and $\eta^{\ast}$ in the proof both take their values in the interval $[0,1]$: the former is the density of $\mu$ with regard to its projection on $\R^d$, while the latter is a compactly supported uniformly continuous approximation to $\eta$ in the $L^2$-norm. Therefore, the function $(\eta(X)-\eta^{\ast}(X))^2$ takes values in $[0,1]$ as well. Thanks to (i$^\prime$), the required inequality holds approximately, to within any wanted error, if $n$ is large enough. Thus, in the first displayed formula on top of page 100 one can replace the upper bound of $3\e(1+1+c)$ with, for example, $4\e(1+1+c)$, provided $n$ is large enough. This will do just as well. The rest of the proof remains unchanged.
\end{pot1}

\subsection{Proof of Theorem \ref{t:kann}} We apply Theorem \ref{t:stone2} with the weights $W_{n,i}$ defined as follows: $W_{n,i}(x)=1/k$ if $X_i$ is among the $(k,c)$-approximate nearest neighbours of $x$ as returned by the oracle (the indexing scheme), and $0$ otherwise. The interpretation of the expected value will depend on whether the indexing scheme is assumed to be randomized or deterministic, however this does not affect the proof.

Clearly, the weights are non-negative. Since for any given $x$ there are precisely $k$ $(k,c)$-approximate nearest neighbours returned, all but $k$ weights vanish at the point $x$, and the weights add up to one almost surely. 

The condition $(ii)$ follows from a classical observation of Cover and Hart \cite{CH} that in every separable metric space equipped with a Borel probability measure, the $1$-Lipschitz function $\ve_{k\mbox{\tiny -NN}}$ (the smallest radius of a ball containing $k$ nearest neighbours among $n$ datapoints) will converge to zero almost surely provided $k/n\to 0$.

The condition $(iii)$ follows from the definition of weights and the assumption $k\to\infty$. 

It remains to verify the condition $(i)$. 
Denote for $a>0$ and $x\in\R^d$
\[\ve_a(x) =\inf\left\{\e>0\colon \mu\left(B_{\e}(x)\right)\geq a\right\}.\]
Now let $f=\partial\mu/\partial\lambda$ be the density, that is, the Radon--Nikod\'ym derivative of the underlying measure on $\R^d$ with regard to the Lebesgue measure. By the Lebesgue differentiation theorem,
\begin{eqnarray*}
\frac{\mu\left(B_{\e_a(1+c)}(x) \right)}{a} &=& 
\frac{\mu\left(B_{\e_a(1+c)}(x) \right)}{\mu\left(B_{\e_a}(x)\right)} \\
&=& \frac{\mu\left(B_{\e_a(1+c)}(x) \right)}{\lambda\left(B_{\e_a(1+c)}(x) \right)}\cdot\frac{\lambda\left(B_{\e_a(1+c)}(x) \right)}{\lambda\left(B_{\e_a}(x) \right)}\cdot \frac{\lambda\left(B_{\e_a}(x) \right)}{\mu\left(B_{\e_a}(x) \right)} \\
&\overset{a\to 0}\longrightarrow & f(x)\cdot (1+c)^d\cdot f(x)^{-1} \\
&=& (1+c)^d,
\end{eqnarray*}
where the convergence is $\mu$-almost surely and, since it is clearly dominated, also in probability.
For $n$ suitably large, $\mu\left(B_{\e_a(1+c)}(x) \right)\leq 2a(1+c)^d$ for all $x$ except for a set of measure $\rho=\rho(n)\to 0$ as $n\to\infty$.

Next we apply the uniform Glivenko--Cantelli theorem to estimate the number of datapoints in $B_{\e_a(1+c)}(x)$.
The VC dimension of the family of all Euclidean balls in $\R^d$ is $d+1$ \cite{dudley:73}. As a consequence, for any $\e>0$, $\delta>0$, if
\[n\geq  \max\left\{\frac{8(d+1)}{\e}\lg\frac{8e}{\e},\frac 4{\e}\lg\frac{2}{\delta} \right\},
\]
then with confidence $1-\delta$ the $\mu$-measure and the empirical measure of every ball differ between themselves by less than $\e$ (\cite{vidyasagar}, Theorem 7.8). Since for $\e=\log n/n$ the expression on the right hand side is of the order $n- n\log\log n/\log n <n$, it follows that for $\e=\omega(\log n/n)$ and a fixed $\delta>0$ the conclusion follows for $n$ sufficiently large. Due to our assumption on $k$, we can set $\e = \Theta(k/n)$ and conclude that, 
again for $n$ sufficiently large, with high confidence, as $n\to\infty$, we have
\[\left\vert X\cap B_{\e_{2k/n}(1+c)}\right\vert \leq 6k(1+c)^d,
\]
and besides 
\[\left\vert X\cap B_{2k/n}(x)\right\vert\geq k.\]
Thus, with confidence $1-\delta$, if $X$ is among $k$-ANN of $X^\prime$, then
the empirical measure $\mu_n$ of the ball of radius $\norm{X-X^\prime}$ centred at $X^\prime$ is at most $6k(1+c)^d$. 

According to Lemma 1.11 of \cite{DGL}, page 171 (Stone's Lemma), the empirical measure of the set of all such $X^\prime$ is at most $\gamma_d\cdot 6k(1+c)^d= 6k(1+c)^\gamma_d$, where $\gamma_d$ is an absolute constant only depending on the dimension $d$. This means that for all samples $X_1,X_2,\ldots,X_n$ of measure $1-\delta$ and a random $X$ independent of $X_i$'s the number of points $X_i$ having $X$ as their $k$-ANN is bounded by
\[6k(1+c)^\gamma_d.\]
Denote the set of i.i.d. samples verifying this condition by $G\subseteq \Omega^n$. One has $\mu(G)\geq 1-\delta$. According to the ``confidence is cheap'' principle, one can assume here that $\delta\to 0$ with any rate of convergence subexponential in $n$, for instance, as $1/n$. 

Now we proceed to verifying the condition $(i)$. Let $A\subseteq\R^d$ be a Borel subset. The quantity that we need to bound can be estimated as follows:
\begin{eqnarray*}
\E\left\{\sum_{i=1}^n W_{n,i}(x)\chi_A(X_i)\right\} &\leq & 
\frac 1k \E\left\{\sum_{i=1}^n I_{\{X_i\mbox{\tiny can be among }k \mbox{\tiny ANN of }X\}}\chi_A(X_i)\right\}\end{eqnarray*}
\[= \frac 1k  \E\left\{\chi_A(X)\sum_{i=1}^n I_{\{X\mbox{\tiny can be among }k \mbox{\tiny ANN of }X_i \mbox{ in }X_1,\ldots,X_{i-1},X,X_{i+1},\ldots,X_n\}}\right\}\] 
One has
\begin{eqnarray*}
&&\frac 1k  \E\left\{\chi_G\chi_A(X)\sum_{i=1}^n I_{\{X\mbox{\tiny can be among }k \mbox{\tiny ANN of }X_i \mbox{ in }X_1,\ldots,X_{i-1},X,X_{i+1},\ldots,X_n\}}\right\} \\
&&\leq \frac 1k \E\left\{\chi_A 6k(1+c)^\gamma_d \right\} 
= 6\mu(A)(1+c)^\gamma_d
\end{eqnarray*}
and
\begin{eqnarray*}
&&\frac 1k  \E\left\{\chi_{\Omega\setminus G}\chi_A(X)\sum_{i=1}^n I_{\{X\mbox{\tiny can be among }k \mbox{\tiny ANN of }X_i \mbox{ in }X_1,\ldots,X_{i-1},X,X_{i+1},\ldots,X_n\}}\right\} \\
&&\leq \frac 1k \cdot \frac 1n \cdot \mu(A)\cdot n = \frac{\mu(A)}k.
\end{eqnarray*}

This finishes the proof.
\qed

\section{Query instability in high dimensions} 

The $k$-ANN classifier has not been tested in practice.
However, within a theoretical model, it is still not free from the curse of dimensionality. Namely, assuming the datapoints follow a high-dimensional distribution on $\R^d$ (such as the gaussian), it is not difficult to prove, as a version of the well-known ``empty space paradox,'' that for a fixed $c>0$ in the limit $d\to\infty$ the number of datapoints must grow exponentially in dimension $d$ in order to maintain the consistency of the algorithm. Indeed, the ball of radius $(1+c)\ve_{\mbox{\tiny $k$-NN}}$ around the query point will contain $\gg k$ datapoints, so it is conceivable that the labels of $k$ points chosen among them by the oracle will be highly biased.

Here are two examples. 
\begin{figure}[ht]
\begin{center}
  \scalebox{0.45}[0.45]{\includegraphics{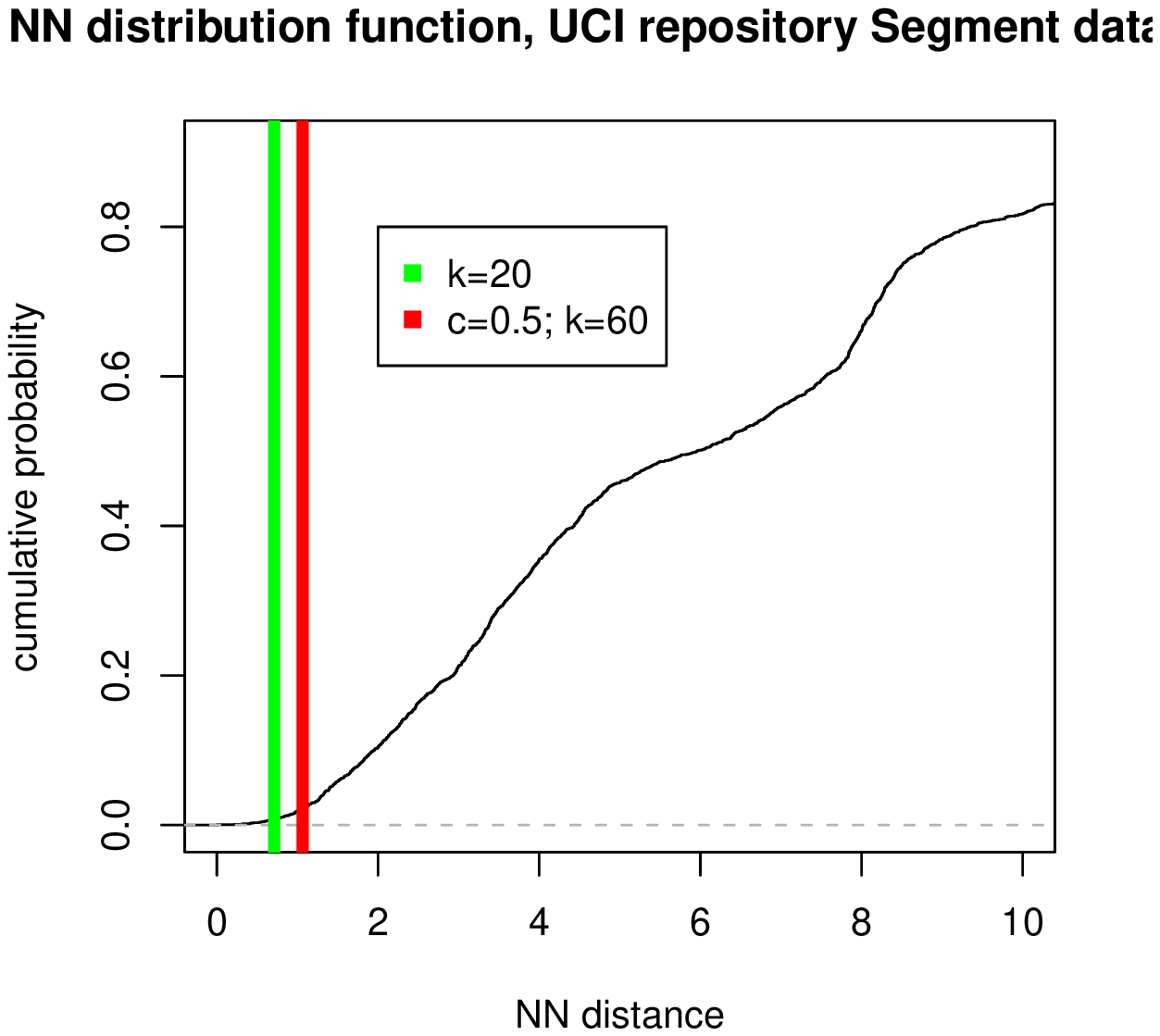}} 
  \hskip .5cm
  \scalebox{0.45}[0.45]{\includegraphics{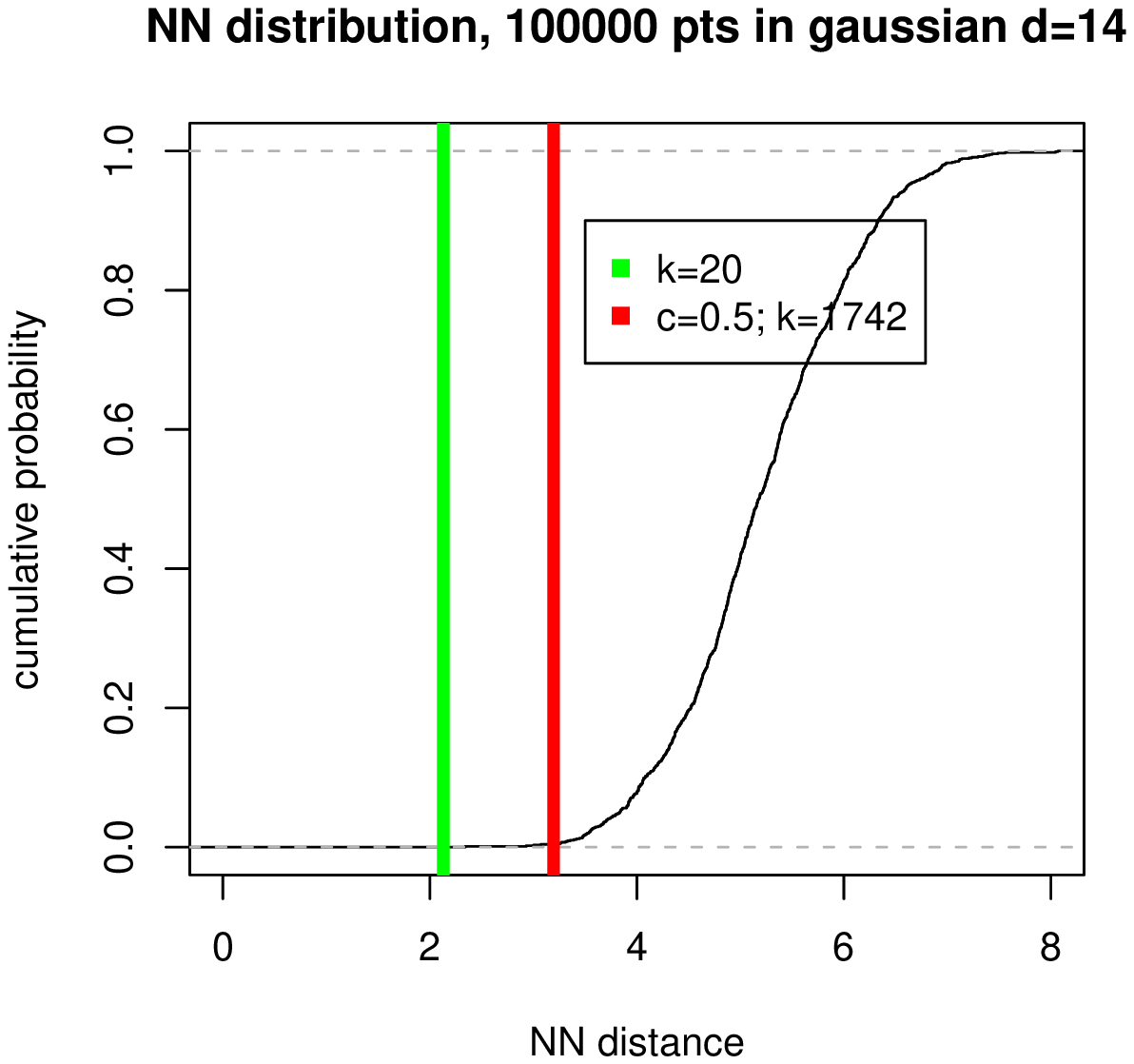}}
  \caption{Average empirical measure of the balls of radius $(1+c)\ve_{\mbox{\tiny $k$-NN}}$.}
  \label{fig:instability}
\end{center}
\end{figure}
The first one is the Segment dataset of the UCI data repository, which has a relatively low intrinsic dimension in any possible sense. The second is a randomly drawn dataset from the gaussian distribution in dimension $14$, whose instrinsic dimension can be described as medium. The graph of the distribution function of the average number of nearest neighbours depending on the distance to the query point is shown in black. Set $k=20$ and $ c=0.5$.
The left vertical line corresponds to the average value of the $k$-NN radius $\ve_{\mbox{\tiny $k$-NN}}$, and the second line corresponds to $(1+c)\ve_{\mbox{\tiny $k$-NN}}$. For the Segment dataset, the latter ball contains on average $60$ datapoints. However, for the gaussian the corresponding value is already $1742$.

\begin{figure}[ht]
\begin{center}
  \scalebox{0.25}[0.25]{\includegraphics{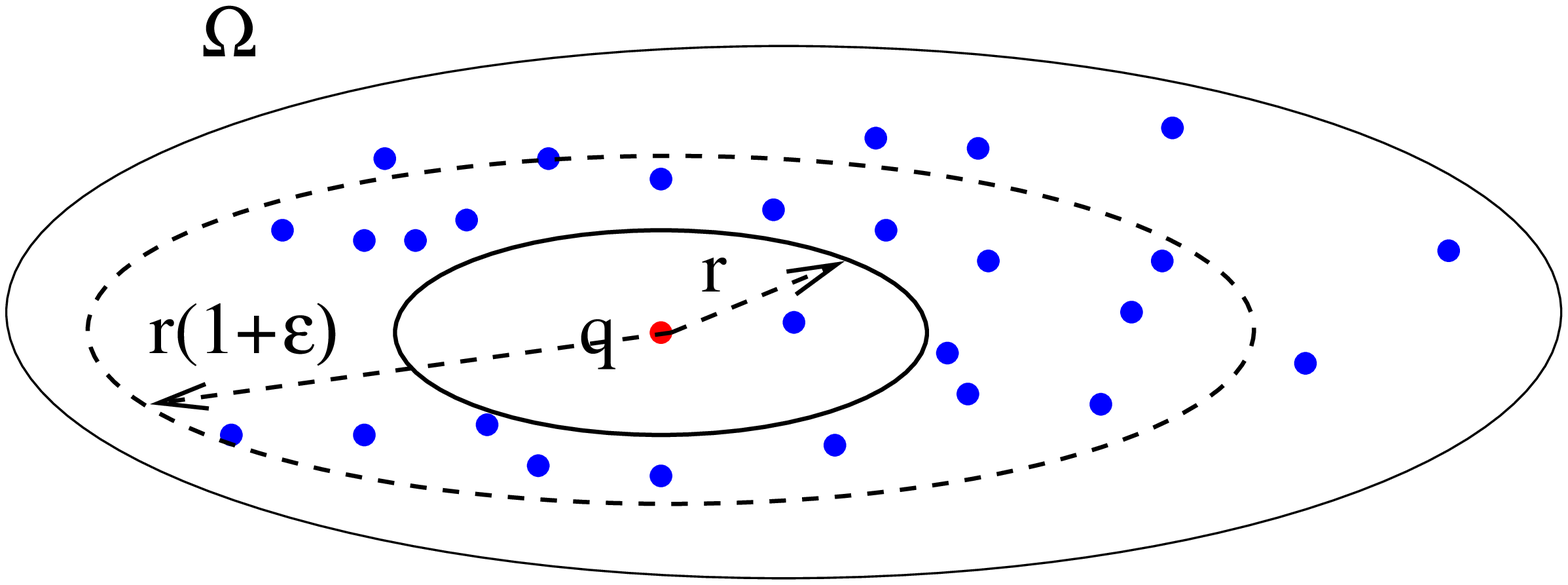}} 
  \caption{Query instability.}
  \label{fig:unstable}
\end{center}
\end{figure}

This brings us to the following definition.
Let us say, following \cite{BGRS}, that a range query $(q,r)$ is {\em $c$-unstable}  if the $(1+c)\ve_{NN}(q)$-ball around $q$ contains at least a half of all datapoints. (Cf. Figure \ref{fig:unstable}.)

Under the subexponential data size growth assumption
\[d=\omega(\log n),\]  
as well as a certain general assumption of intrinsic high-dimensionality of the underlying measure distribution \cite{pestov:07,pestov:08}, one can prove that asymptotically an overwhelming majority of queries are $(1+c)$-unstable. (Cf. theorem 2.1 in \cite{pestov:12}.) This assumption is met by the gaussian measures on $\R^d$, the uniform measures on the cubes $\I^d$, the uniform measures on the Hamming cubes $\{0,1\}^d$, and so forth.
In such a situation, the $(k,c)$-ANN search problem can be essentially answered by returning $k$ randomly picked datapoints. The $k$-ANN classifier becomes meaningless, because the Bayes error approaches $1/2$ in the limit $d\to\infty$.

The exponential rate of growth of dataset size $n$ with regard to dimension $d$ is of course unrealistic. This means that at least in some theoretical situations (i.i.d. sampling from an artificial high-dimensional distribution) even allowing for approximate nearest neighbours will not save the $k$-NN classifier from the curse of dimensionality.

\section{Borel dimensionality reduction}
Basic concepts of descriptive set theory \cite{kechris} offer a new approach to dimensionality reduction in the context of statistical learning. With this purpose, let us re-examine the standard setting for (non-parametric) statistical classification as outlined in Section \ref{s:stone} above.

\subsection{Borel sets, mappings, and isomorphisms\label{ss:borel}}
Recall that a family $\mathscr A$ of subsets of a set $X$ is a {\em sigma-algebra} if $\mathscr A$ contains $X$ and is closed under the complements and unions of countable subfamilies. A set $X$ equipped with a sigma-algebra $\mathscr A$ of subsets is called a {\em measurable space}. 
Let now $X$ be a separable metric space. The {\em Borel sigma-algebra} of $X$, which we will denote ${\mathscr B}_X$, is the smallest sigma-algebra of subsets of $X$ containing all open balls. In particular, ${\mathscr B}_X$ contains all open and all closed subsets of $X$, all intersections of countable families of open sets ($G_\delta$-sets), all unions of countable families of closed sets ($F_\sigma$-sets), and so on. In fact, Borel subsets are so numerous that it is not easy to exhibit a constructive example of a non-Borel subset of a separable metric space such as $\R$. 

A mapping $f\colon X\to Y$ between two separable metric spaces is {\em Borel} (or {\em Borel measurable}) if the inverse image $f^{-1}(B)$ of each Borel subset of $Y$ is Borel in $X$. Equivalently, the inverse image of every open ball in $Y$ is a Borel subset of $X$. For instance, the indicator function of the rationals is a Borel function. By changing the values of a Lebesgue measurable function on a suitable null-set, one can obtain a Borel function. This stresses how numerous Borel sets and functions are.

A bijective Borel mapping whose inverse mapping is also Borel is called a {\em Borel isomorphism}. It turns out that from the Borel isomorphic viewpoint, metric spaces do not differ between themselves that much. 
More precisely, two complete metric spaces $X$ and $Y$ of the same cardinality are Borel isomorphic. 
Thus, the Cantor set, the closed unit interval, the Euclidean space $\R^d$, the infinite-dimensional separable Hilbert space $\ell^2$, and in fact all separable Fr\'echet spaces different from zero space are all pairwise Borel isomorphic between themselves. Their Borel structure is that of a {\em standard Borel space} of cardinality continuum.

An example of a Borel isomorphism between the interval $[0,1]$ and the square $[0,1]^2$ can be obtained by interlacing between themselves the binary expansions of $x$ and $y$ of a pair $(x,y)\in\I^2$ (subject to the usual precautions concerning infinite strings of ones):
\begin{equation}
  \label{eq:borelmap}
  [0,1]^2\ni(0.a_1a_2\ldots, 0.b_1b_2\ldots)\mapsto (0.a_1b_1a_2b_2\ldots)\in [0,1].\end{equation}

\begin{figure}[ht]
\begin{center}
  \scalebox{0.25}[0.25]{\includegraphics{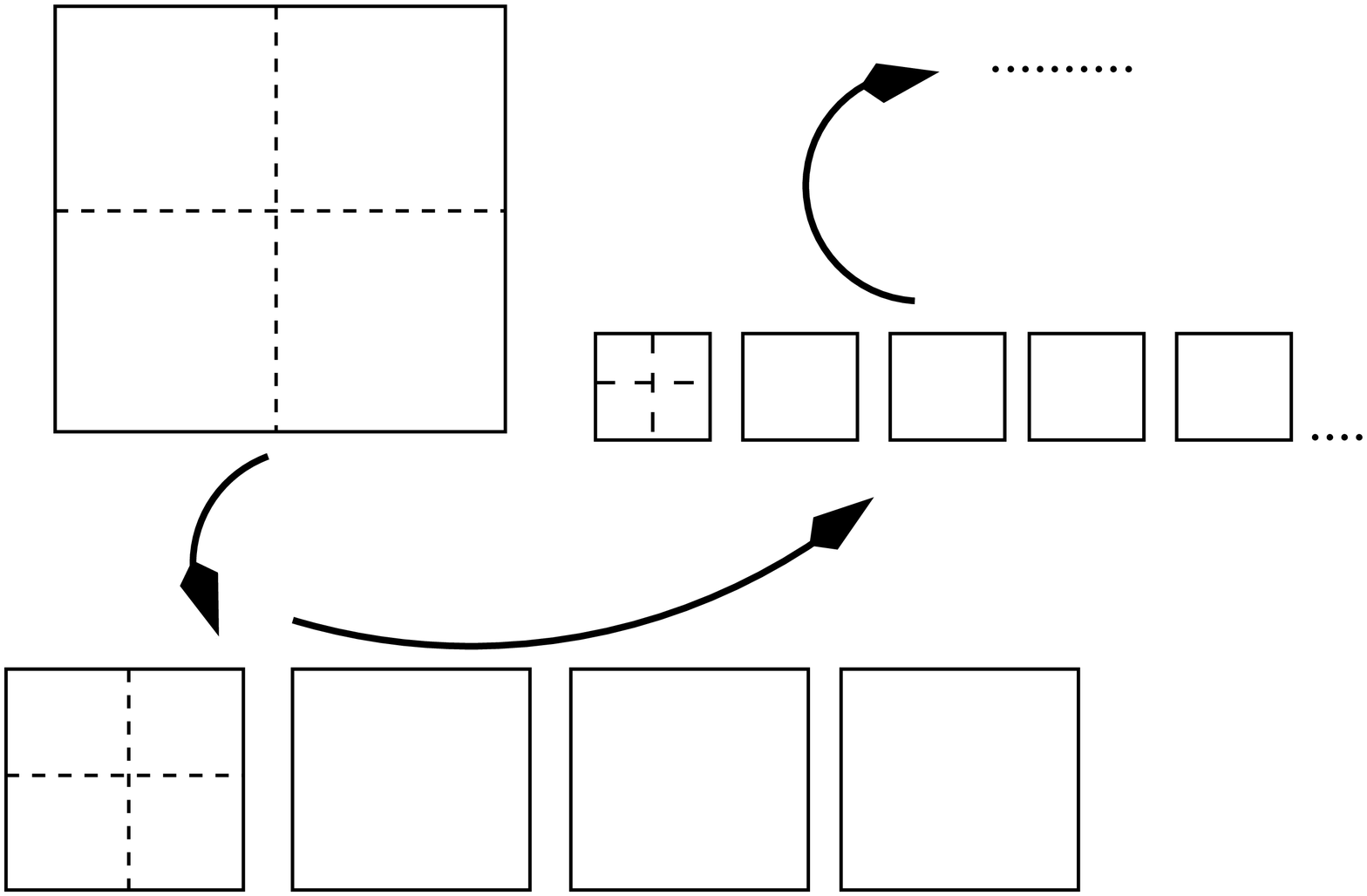}} 
  \caption{Constructing a Borel isomorphism between the square and the interval.}
  \label{fig:borel_isomorphism}
\end{center}
\end{figure}

A geometric representation of this isomorphism can be seen in Figure \ref{fig:borel_isomorphism}.

This of course extends to any number of dimensions.

Usually the mappings performing the data reduction of the domain are assumed to be continuous, even Lipschitz. However, if one looks at the existing theoretical model laying down a foundation for statistical learning, one can notice that Stone's theorem is in fact insensitive to the {\em Euclidean structure} (that is, either metric or topological structure) on the domain as long as the underlying {\em Borel structure} remains intact. This allows for a very simple ``Borel isomorphic data reduction'' to the one-dimensional case, after which the 
algorithm still remains universally consistent. 

\subsection{Borel isomorphic dimension reduction}
The following result, although straightforward, offers, in our opinion, a potentially interesting new approach to dimensionality reduction in statistical learning theory. 
We consider it as the central result of the work reported. 

Recall that a {\em standard Borel space} is a complete separable metric space equipped with its Borel structure.

\begin{theorem}
\label{th:main}
Let $\Omega$ be a standard Borel space. Fix a Borel isomorphism $\phi\colon\Omega\to X$, where $X=(X,d_X)$ is a metric space in which the $k$-NN learnig rule is universally consistent (for instance, $\R^m$ or its metric subspace). Define a metric $\rho$ on $\Omega$ by 
\[\rho(x,y) = d_X(\phi(x),\phi(y)).\]
Then the learning rule on $\Omega$ given by the $k$-NN rule with regard to the metric $\rho$ is universally consistent.
\end{theorem}

Before proving the result, we need to fix notation and terminology.

A {\em metric space with measure,} or an $mm${\em -spaces,} is a triple $(\Omega,\rho,\mu)$, consisting of a separable metric space $(\Omega,\rho)$ and a Borel probability measure $\mu$ on this space. This is an important notion in 
modern geometry and functional analysis \cite{Gr}.  

The basic object of classification theory will be very similar, with the only difference that the probability measure $\mu$ is now defined on $\Omega\times\{0,1\}$. Equivalently, such an object can be described as a metric space $(\Omega,\rho)$ equipped with a pair of finite measures $\mu_0,\mu_1$, whose total mass adds up to one: $\mu_0$ is the restriction of $\mu$ to $\Omega\times\{0\}$, and $\mu_1$, to $\Omega\times\{1\}$. Let us call such objects $mm2${\em -spaces}.

Two metric spaces with measure $(\Omega_i,\rho_i,\mu_i)$, $i=1,2$ are {\em isomorphic} if there is an {\em isomorphism} between them, that is, a mapping $\phi\colon \Omega_1\to\Omega_2$ which is an isomorphism of measure spaces and which preserves the metric almost everywhere. The concept of an isomorphism between our $mm2$-spaces is defined similarly: it is a measurable mapping $\phi$ which preserves $\mu_0$, $\mu_1$, and which preserves pairwise distances between points $(\mu_0+\mu_1)$-almost everywhere.

Let $(\Omega,\rho,\mu_0,\mu_1)$ and $(\Omega^\prime,\rho^\prime,\mu_0^\prime,\mu_1^\prime)$ be two isomorphic $mm2$-spaces, with an isomorphism $\phi\colon \Omega\to\Omega^\prime$.
Let $\mathcal L$ be a learning rule in $\Omega$. Then one can define a learning rule in the space $(\Omega^\prime,\rho^\prime,\mu_0^\prime,\mu_1^\prime)$ using the isomorphism, as follows. Denote $\phi^{-1}$ the inverse measurable isomorphism to $\phi$. We will denote by the same symbol $\phi^{-1}(s)$ the image of a labelled $n$-sample $s=(x_1,x_2,\ldots,x_n,y_1,y_2,\ldots,y_n)$, that is,
\[\phi^{-1}(s) = (\phi^{-1}(x_1),\phi^{-1}(x_2),\ldots,\phi^{-1}(x_n),y_1,y_2,\ldots,y_n).\]
Now set
\[{\mathcal L}^{\phi}(s) = {\mathcal L}_n(\phi^{-1}(s))\circ \phi.\]
This ${\mathcal L}^{\phi}$ is a learning rule on $\Omega^\prime$, a {\em transport} of the rule $\mathcal L$ along the map $\phi$.
The following should now be obvious.

\begin{lemma}
\label{l:error}
The learning rule ${\mathcal L}^\phi$ in $\Omega^\prime$ has the same learning error as $\mathcal L$ does in $\Omega$.
\qed
\end{lemma}

\begin{pot2}
  It is enough to notice that the $k$-NN classifier in the metric space $(\Omega,\rho)$ is the transport along $\phi$ of the $k$-NN classifier in the metric space $(X,d_X)$. For every probability distribution $\mu$ on $\Omega\times\{0,1\}$, denote $\phi_\ast\mu$ the push-forward of $\mu$ along $\phi$. This is a Borel probability distribution on $X\times\{0,1\}$, and clearly the Bayes error of $\mu$ equals that of $\phi_\ast\mu$.
  In view of our hypothesis of the universal consistency of the $k$-NN classifier in $X$, the learning error of the classifier equals to Bayes error. Due to Lemma \ref{l:error}, the same conclusion holds for the $k$-NN classifier in the metric space $(\Omega,\rho)$. 
\end{pot2}

In particular, there is always a Borel isomorphic reduction of the problem in $\R^d$ to $d=1$. Moreover, the reduction even applies to functional data learning in the most general situation imaginable, when $\Omega$ is an arbitrary separable metric space, for instance a separable Banach space, or even a separable Fr\'echet space. 

\begin{example}
  The histogram learning rule in the cube $\I^d$ is a Borel isomorphic reduction to the Cantor set (a zero-dimensional compact metrizable space without isolated points) equipped with a non-archimedian metric.
\end{example}

\begin{example}
The distance metric learning methods, such as LMNN (see e.g. \cite{WSS}), are based on selecting an alternative euclidean metric in $\R^d$. This is equivalent to selecting a linear isomorphism $\phi$ from $\R^d$ to itself and using the learning rule ${\mathcal L}^\phi$ in the original space. Here, $\phi$ is of course not just a Borel isomorphism, but moreover a homeomorphism.
\end{example}

\begin{table}
  \begin{center}
    \begin{tabular}{|l|c|c|c|c|c|c|c|c|}
  \hline
  & \multicolumn{2}{|c|}{Iris} 
  & \multicolumn{2}{|c|}{Diabetes}
  & \multicolumn{2}{|c|}{Ionosphere}
  & \multicolumn{2}{|c|}{Balance-scale}  \\ 
  \hline
Dimension reduction   &\multicolumn{2}{|c|}{4 $\longrightarrow$ 1}   
&\multicolumn{2}{|c|}{8 $\longrightarrow$ 1}  
&\multicolumn{2}{|c|}{33 $\longrightarrow$ 1}  
& \multicolumn{2}{|c|}{4 $\longrightarrow$ 1} \\ 
 Number of Instances      &\multicolumn{2}{|c|}{150}  
&\multicolumn{2}{|c|}{768}  
&\multicolumn{2}{|c|}{351} 
&  \multicolumn{2}{|c|}{625} \\
\cline{2-9}
Correctly Classified &144 & 140 &569 & 562& 316& 300 & 561 & 407        \\
Same, \%                            &96.0 & 93.3 &74.1 & 73.0 & 90.0 &85.5 & 89.9 &65.1\\
Incorrectly Classified &6 & 10 &  63 &199 & 35 &51 & 207 & 218   \\
Same, \%                              &4.0 & 6.67 & 25.9 &27.0 & 10.0 &14.5 &10.8&34.9 \\
\cline{2-9}
Optimal value of $k$ & 6 & 3 & 17 & 10 & 2 &8 & 9& 18 \\ 
\hline
\end{tabular}
\caption{$k$-NN classification of some datasets in UCI repository before and after a Borel dimensionality reduction as in Eq. (\ref{eq:borelmap}), using RWeka ($1\leq k\leq 20$, $10$-fold cross-validation).}
\label{t:uci}
\end{center}
\end{table}

\section{Discussion}

In this article, we suggested two novel approaches to dimensionality reduction in the context of the $k$-NN classification: the $k$-approximate nearest neighbour rule and the Borel isomorphic dimensionality reduction. The closest counterpart in the literature is an approach based on a combination of the $k$-NN classifier with random projections \cite{HDWB,FJS}. Notice, however, that this is different from our approaches: first, not every indexing scheme for ANN search is based on random projections, and second, projections are not Borel isomorphisms. 

A more technical paper about the approximate $k$ nearest neighbour classifier in which the algorithm has been implemented and tested is currently in preparation \cite{DPS}. The Borel isomorphic dimensionality reduction is easy to implement, and we invite the readers to try their hand at it. As a word of caution, if the original domain is high dimensional, this may necessitate using floating-point arithmetic.

The initial experiments with data from the UCI machine learning repository (Table \ref{t:uci}) show that the 
Borel isomorphic reduction succeeds at least for some datasets, and fails for others. On the one hand, the richness of the class of Borel isomorphisms is enormous, and the failure can be always attributed to a poor choice of a reduction. On the other hand, it is definitely hard to expect such a simple idea to give a panacea of the curse of dimensionality. What is probably a realistic expectation, is a possibility to slash a high dimension by a given factor without degraing the performance, by arranging the dimensions of a domain in groups of $k\ll n$ and performing a Borel isomorphic reduction $\R^k\to\R$ on every such group separately. An interesting perspective is a theory of capacity for {\em families} of Borel isomorphisms between a given domain and a fixed lower-dimensional space (e.g. $\R$), enabling search for an optimal Borel data reduction for a given dataset. 

At a theoretical level, this brings up the question, 
what is dimension in the context of statistical learning?
It took mathematicians roughly half a century, to isolate the correct notion of dimension of a topological space and obtain the basic results of the theory (roughly, 1873--1921, see \cite{crilly}). Will it take the same amount of time to put forward a satisfactory theory of intrinsic dimension of data in the context of statistical learning, in order to explain away the curse of dimensionality?
 
What our investigation demonstrates, is that such a notion should reflect not the dimension of the domain per se, but rather the complexity of the target classifier in the setting of a $mm2$-space consisting of a metric domain $(\Omega,\rho)$ and a probability distribution on $\Omega\times\{0,1\}$. An important factor is the isoperimetric behaviour of the unknown target concept $C$, by which we mean the rate of growth of the function
\[\e\mapsto \mu(C_\e),\]
where $C_\e$ is the $\e$-neighbourhood of the concept. A fast growing isomerimetric function reflects the high complexity of the margin and a consequent difficulty of finding a classifier. 

\subsection*{Acknowledgements} 
This is an extended write-up of the talk of the same title given by the author on August 25, 2011 at the Dagstuhl Perspective Seminar 11341 ``Learning in the context of very high dimensions'' (organized by Michael Biehl, Barbara Hammer, Erzs\'ebet Mer\'enyi, Alessandro Sperduti, and Thomas Villmann). The original slides of the talk are available on the web page \\
{\tt http://www.dagstuhl.de/mat/index.en.phtml?11341}

Among many interesting talks at the Dagstuhl seminar, I want to particularly single out those by Fabrice Rossi, Udo Seiffert, and Kilian Weinberger, from which I have learned a number of important results and references, including \cite{CG}, \cite{hughes:68}, and \cite{WSS}, correspondingly. Stimulating discussions with Erzs\'ebet Mer\'enyi are also acknowledged.

\end{document}